\documentclass{article} 
\usepackage{iclr2025_conference,times}

\usepackage{amsmath,amsfonts,bm}

\def\eqref#1{equation~\ref{#1}}

\def\1{\bm{1}}

\DeclareMathAlphabet{\mathsfit}{\encodingdefault}{\sfdefault}{m}{sl}
\SetMathAlphabet{\mathsfit}{bold}{\encodingdefault}{\sfdefault}{bx}{n}

\DeclareMathOperator*{\argmin}{arg\,min}

\usepackage{hyperref}
\usepackage{url}
\usepackage{booktabs}       
\usepackage{enumerate}
\usepackage{multirow} 
\usepackage{multicol}
\usepackage{amsmath}
\usepackage{amsthm}
\usepackage{amssymb}
\usepackage{bbm}

\newtheorem{lemma}{Lemma}

\newtheorem{proposition}{Proposition}
\usepackage{graphicx}
\usepackage{subcaption}
\usepackage{wrapfig}
\usepackage{float}
\usepackage{algorithm}
\usepackage{algpseudocode}
\definecolor{royalblue}{rgb}{0.25, 0.41, 0.88}
\definecolor{amaranth}{rgb}{0.9, 0.17, 0.31}

\title{Diverse Policies Recovering via Pointwise \\ Mutual Information Weighted Imitation\\ Learning}

\author{Hanlin Yang$^{12}$\footnotemark[1]\hspace{1.5mm}\footnotemark[2]\hspace{1mm}, Jian Yao$^{2}$\footnotemark[1]\hspace{1mm}, Weiming Liu$^{2}$, Qing Wang$^{2}$, Hanmin Qin$^{2}$, Hansheng Kong$^{2}$,\\ \textbf{Kirk Tang$^{2}$, Jiechao Xiong$^{2}$, Chao Yu$^{1}$\footnotemark[3]\hspace{1mm}, Kai Li$^{3}$, Junliang Xing$^{4}$, Hongwu Chen$^{2}$,}\\ \textbf{Juchao Zhuo$^{2}$, Qiang Fu$^{2}$, Yang Wei$^{2}$, Haobo Fu$^{2}$\footnotemark[3]}}

\footnotetext[1]{Equal contribution}
\footnotetext[2]{Work done during an internship at Tencent AI Lab}
\footnotetext[3]{Equal advising}

\iclrfinalcopy 
\begin{document}

\maketitle

{\small
\vspace*{-10mm}
\hspace*{1mm}
$^{1}$ Sun Yat-sen University, Guangzhou, China\\
\hspace*{1mm}
$^{2}$ Tencent AI Lab, Shenzhen, China\\
\hspace*{1mm}
$^{3}$ Institute of Automation, Chinese Academy of Sciences, Beijing, China\\
\hspace*{1mm}
$^{4}$ Tsinghua University, Beijing, China

\vspace*{-2mm}
\texttt{\hspace*{2mm}yanghlin7@mail2.sysu.edu.cn, yuchao3@mail.sysu.edu.cn, kai.li@ia.ac.cn,\\
\hspace*{2mm}jlxing@tsinghua.edu.cn,  \{nigeljyao, weimingliu, drwang, hanminqin,\\
\hspace*{2mm}hanskong, kirktang, jcxiong, hongwuchen, jojozhuo, leonfu, willyang,\\ \hspace*{2mm}haobofu\}@tencent.com}
}

\vspace*{7mm}
\begin{abstract}
Recovering a spectrum of diverse policies from a set of expert trajectories is an important research topic in imitation learning. After determining a latent style for a trajectory, previous diverse policies recovering methods usually employ a vanilla behavioral cloning learning objective conditioned on the latent style, treating each state-action pair in the trajectory with equal importance. Based on an observation that in many scenarios, behavioral styles are often highly relevant with only a subset of state-action pairs, this paper presents a new principled method in diverse polices recovery. In particular, after inferring or assigning a latent style for a trajectory, we enhance the vanilla behavioral cloning by incorporating a weighting mechanism based on \emph{pointwise mutual information}.
This additional weighting reflects the significance of each state-action pair's contribution to learning the style, thus allowing our method to focus on state-action pairs most representative of that style.
We provide theoretical justifications for our new objective, and extensive empirical evaluations confirm the effectiveness of our method in recovering diverse policies from expert data.
\end{abstract}

\section{Introduction}

Imitation Learning (IL) is about observing expert demonstrations in performing a task and learning to mimic those actions~\citep{hussein2017imitation, osa2018algorithmic}. Vanilla behavioral cloning (BC)~\citep{pomerleau1991efficient} learns a mapping from state to actions using expert state-action pairs via supervised learning, which is simple to implement but may have the issue of compounding errors.
Generative Adversarial Imitation Learning (GAIL)~\citep{ho2016generative} mitigates the issue via learning both a discriminator and a policy. Despite their wide applications, these methods lack mechanisms to generate diverse policies, which may be essential in certain tasks.

There has been an increase in recent research addressing policy diversity in imitation learning~\citep{li2017infogail,wang2017robust,zhan2020learning,shafiullah2022behavior,mao2023stylized}, which can generally divided into two categories. In one category, the latent style $z$ of a trajectory is inferred in an unsupervised manner, for instance, by an expectation maximization (EM) procedure. In the other category, the style $z$ of a trajectory is determined by a user-specified function, for instance, a programmatic labeling function. No matter in which category, samples that are used to train a style-conditioned policy $\pi(a|s,z)$ are treated with equal importance in those methods. However, in \newpage many cases, we observe that the relevance of different state-action pairs to the trajectory style can vary significantly.
For example, in autonomous driving tasks, the diversity of overtaking policies (from the left or right) is primarily relevant to the overtaking period in the trajectory. In other words, the preceding normal driving period is less relevant to the overtaking diversity. Again,
in a basketball game, the diversity of shooting position behaviors is primarily related to the part leading up to the shot, while it is less relevant to other parts in the trajectory. In a nutshell, policies with different styles could have significantly different degrees of overlap in different areas of the state-action space.
Therefore, when learning a style-conditioned policy, the relevance between the state-action pairs in the trajectory and the behavioral style should be taken into consideration. 

In this paper, we propose a new diverse policies recovering method by leveraging the relevance of the state-action pairs with the trajectory styles.
We focus on the situation where the style label of a trajectory has been provided, as unsupervised learning (either by EM or mutual information maximization) of the style $z$ leads to uncontrolled styles and often only qualitative evaluation. In particular, we introduce an additional importance weight based on Pointwise Mutual Information (PMI)~\citep{church1990word, manning1999foundations, bouma2009normalized}, in traditional conditional BC, to quantify the relevance of state-action pairs with the conditioned style. Intuitively, state-action pairs with a larger posterior of the corresponding style $p(z|s,a)$ are given a larger weight. 
In practical implementation, we utilize Mutual Information Neural Estimation (MINE)~\citep{belghazi2018mutual} to estimate the PMI between state-action pairs and style variables. We term the proposed method Behavioral Cloning with Pointwise Mutual Information Weighting (BC-PMI).

Our theoretical analysis indicates that our new weighted learning objective unifies two extreme cases in recovering diverse policies. When the mutual information between the style and the state-action pair is zero, which means there is no style diversity in expert data, our objective degenerates to vanilla BC, which views the data as generated from one policy. By contrast, when policies with different styles have no overlap in the state-action space, our objective degenerates to learning different style polices separately. Empirical results in Circle 2D, Atari games and professional basketball player dataset demonstrate that BC-PMI achieves better performance in recovering diverse policies than the baseline methods.

\section{Related Work}

\paragraph{Imitation Learning}
Imitation learning (IL) methods are designed to mimic the behaviors of experts.
Behavioral Cloning (BC)~\citep{pomerleau1991efficient}, a well-known IL algorithm, learns a policy by directly minimizing the discrepancy between the agent and the expert in the demonstration data.
However, offline learning methods like BC suffer from compounding errors and the inability to handle distributional shifts during evaluation~\citep{ross2011reduction, fujimoto2019off, wu2019behavior, peng2019advantage, kostrikov2021offline}.
Inverse Reinforcement Learning (IRL)~\citep{ng2000algorithms, arora2021survey}, another type of IL, learns a reward function that explains the expert behavior and then uses this reward function to guide the agent's learning process.
Popular IRL approaches like Generative Adversarial Imitation Learning (GAIL)~\citep{ho2016generative} and Adversarial Inverse Reinforcement Learning (AIRL)~\citep{fu2017learning} use adversarial training to learn a policy that is similar to the expert policy while being robust to distributional shift.
However, these methods are limited to imitating a single policy and do not address the issue of promoting diverse policies.
When imitating diverse policies, BC approaches using supervised learning tend to learn an average policy, which does not fully capture the range of diverse behaviors~\citep{codevilla2018end}.
GAIL tends to learn a policy that captures only a subset of the expert's control behaviors, which can be viewed as modes of distribution.
Consequently, the learned policy fails to cover all styles of the expert's diverse behaviors~\cite{wang2017robust}.

\paragraph{Policy Diversity}
Diversity is crucial in imitation learning algorithms, especially in practical control tasks and multi-player games~\citep{zhu2018reinforcement}, as diverse policies in control tasks can enhance the robustness of adapting to various environments. In contrast, AI with diverse policies can maximize the player's experience and the ornamental value in games and competitions~\citep{yannakakis2018artificial}.
Some approaches utilize information-theoretic methods to address this issue and learn the behavioral styles of policies.
InfoGAIL~\citep{li2017infogail} and Intention-GAN~\citep{hausman2017multi} augment the objective of GAIL with the mutual information between generated trajectories and the corresponding latent codes.
\cite{wang2017robust} use a variational autoencoder (VAE) module to encode expert trajectories into a continuous latent variable.
\cite{eysenbach2018diversity} propose a general method called Diversity Is All You Need (DIAYN) for learning diverse skills without explicit reward functions.
DIAYN focuses on discovering skills in online reinforcement learning tasks~\citep{campos2020explore, sharma2019dynamics, achiam2018variational}, whereas our method specifically considers pure offline imitation learning scenarios.
More recently, \cite{mao2023stylized} introduced Stylized Offline RL (SORL), which utilizes unsupervised learning methods to cluster styles and fits policies to each cluster separately, optimizing an equal number of policies while using the KL divergence as a constraint.
However, these methods focus on inferring the latent style or clustering the trajectories with different styles without considering the relevance between the state-action pairs in the trajectory and the behavioral style.
In contrast, we focus on the situation where the style label of a trajectory has been provided.
By introducing the PMI weights, we quantify the relevance of state-action pairs with the conditioned style, allowing the policy to focus on the samples that are highly relevant to the style.

\section{Preliminary}
\subsection{Problem Setting and Vanilla Behavior Cloning}
\label{setup}
We consider the standard Markov Decision Process (MDP)~\citep{sutton2018reinforcement} as the mathematical framework for modeling sequential decision-making problems, which is defined by a tuple $\left \langle \mathcal{S}, \mathcal{A}, P, r, d_0, T \right \rangle$, where $\mathcal{S}$ is a finite set of states, $\mathcal{A}$ is a finite set of actions, $P: \mathcal{S} \times \mathcal{A} \times \mathcal{S} \to \mathbb{R}$ is the transition probability function, $r: \mathcal{S} \to \mathbb{R}$ is the reward function, $d_0$ is the initial distribution, and $T$ is an episode horizon.
A policy $\pi: \mathcal{S} \times \mathcal{A} \to [0,1]$ maps from state to distribution over actions.
Let $d^\pi_t$ and $d^\pi=\frac{1}{T}\sum_{t=1}^T d^\pi_t$ denote the distribution over states at time step $t$ and the average distribution over $T$ time steps induced by $\pi$, respectively.
The vanilla BC loss function is:
\begin{equation}
    \mathcal{L}_{\mathrm{BC}} = \mathop{\mathbb{E}}\limits_{(s,a)\sim \mathcal{D}_e}\big[ -\log\pi(a|s)  \big].
    \label{eq_cbc}
\end{equation}
which aims to maximize the probability of selecting action $a$ for a given policy under the state $s$.



Building upon the basic setup, we focus on an assumption where the expert demonstrations $\mathcal{D}_e$, which consist of many diverse trajectories $\tau$, are collected by stylized expert policies denoted as $\{\pi_e^{(1)}, \pi_e^{(2)},\dots,\pi_e^{(K)}\}$.
Let $z\in\mathcal{Z}$ denotes the variable indicating which stylized policy $\tau$ belongs to, and $p(\tau|z=i)$ denotes the probability of $\tau$ sampled under policy $\pi_e^{(i)}$.
Our objective is to learn a conditioned policy $\pi(a | s,z)$ such that trajectories generated by $\pi(a | s,z)$ closely match the demonstrations in $\mathcal{D}_e$ that exhibit the corresponding style $z$.

\subsection{Mutual Information Neural Estimation}
Mutual Information Neural Estimation (MINE) is a powerful technique for estimating the mutual information between two random variables using neural networks \citep{belghazi2018mutual}. 
It has been widely used in various domains, including representation learning \citep{hjelm2018learning}, generative modeling \citep{chen2016infogan}, and imitation learning \citep{eysenbach2018diversity}.
The key idea behind MINE is to formulate the estimation of mutual information as an optimization problem.
Given two random variables $X$ and $Y$, the mutual information $I(X;Y)$ can be expressed as the Kullback-Leibler (KL) divergence between the joint distribution $P_{XY}$ and the product of marginal distributions $P_X \otimes P_Y$:
\begin{equation}
I(X;Y) = D_{KL}(P_{XY} || P_X \otimes P_Y).
\end{equation}
MINE approximates this KL divergence using a lower bound based on the Donsker-Varadhan representation \citep{donsker1983asymptotic}:
\begin{equation}
I(X;Y) \geq \sup_{T_\theta \in \mathcal{F}} \mathbb{E}_{P_{XY}}[T_\theta] - \log(\mathbb{E}_{P_X \otimes P_Y}[e^{T_\theta}]),
\end{equation}
where $\mathcal{F}$ is a class of functions $T: \mathcal{X} \times \mathcal{Y} \to \mathbb{R}$.
In MINE, this function class is parameterized by a neural network $T_\theta$, which takes as input samples from the joint distribution $P_{XY}$ and the product of marginal distributions $P_X \otimes P_Y$.
The network is trained to maximize the lower bound, equivalent to estimating the mutual information.

\section{Behavioral Cloning with Pointwise Mutual Information Weighting}

As discussed in the introduction, in many real-world applications, the impact of different state-action pairs on the style can vary greatly, and often only a part of the trajectory is highly relevant to the style. Hence, when training a style-conditioned policy, we would like to assign different weights to different state-action pairs based on their relevance to that style. We aim to develop such a method that can capture the specific influence of each $(s,a)$ pair on the style, thereby achieving a more generalized imitation objective:
\begin{equation}
    \min_\pi \mathop{\mathbb{E}}\limits_{(s,a,z)\sim \mathcal{D}_e}\left[ -\log\pi(a|s,z)\cdot \sigma(s,a,z)  \right],
    \label{eq_wbc}
\end{equation}
where $\sigma(s,a,z)$ is a weighting function. Intuitively, state-action pairs that are more exclusive to a style should have larger weights in learning that style-conditioned policy, and vice-versa. Drawing inspiration from the information theory, we introduce the PMI~\citep{church1990word, manning1999foundations, bouma2009normalized} to quantify the contribution of $(s,a)$ when learning a style-conditioned policy:
\begin{equation}
    \mathcal{P}(z;s,a)=\log\frac{p(s,a,z)}{p(s,a)p(z)}=\log\frac{p(z|s,a)}{p(z)}.
\end{equation}

We utilize the exponential of $\mathcal{P}(z;s,a)$ as the weight, which is the ratio of the posterior given $(s,a)$ and the prior of the style $z$.
This weight has the following properties.
When the posterior of style $z$ given $(s,a)$ is larger than the prior, i.e., $p(z|s,a) > p(z)$, this means there is a high relevance between $(s,a)$ and style $z$, and we should give a larger weight.
If $(s,a)$ and style $z$ are nearly independent, then $p(z|s,a) \approx p(z)$.
When we have $p(z|s,a) < p(z)$, this means (s,a) is more likely generated by other styles, and we should give a lower weight here. Accordingly, Eq.(\ref{eq_wbc}) turns into the following form:
\begin{equation}
    \mathcal{L}_{\mathrm{BC-PMI}}(\theta)= \mathop{\mathbb{E}}\limits_{(s,a,z)\sim \mathcal{D}_e}\left[ -\log\pi_{\theta}(a|s,z)\cdot e^{\mathcal{P}(z;s,a)}  \right]=
    \mathop{\mathbb{E}}\limits_{(s,a,z)\sim \mathcal{D}_e}\left[ -\log\pi_{\theta}(a|s,z)\cdot \frac{p(z|s,a)}{p(z)}  \right]
    \label{eq_wwbc}
\end{equation}

\subsection{Theoretical analysis}

We give some theoretical justifications for our BC-MI objective in Eq.(\ref{eq_wwbc}). Considering an extreme case where the mutual information between $(s, a)$ and style $z$ approaches zero, which means policies with different styles have nearly the same state-action distribution.
In this case, the BC-PMI objective degenerates into vanilla BC, which means there is no difference in training style conditioned policies and an average unconditioned policy.
In the opposite extreme case, where all $(s, a)$ pairs exhibit significant style differences, which means trajectories of different styles have minimal overlap, we find that the BC-PMI objective degenerates into behavior cloning on each style.
Formally, we have the following proposition:
\begin{proposition}
(a). When the mutual information $I(Z;S,A)$ equals to $0$, it indicates that there is no distinction in the trajectory style corresponding to all the state-action pairs.
In this case, the BC-PMI objective degenerates to the vanilla behavior cloning objective:
\begin{equation}
\argmin_{\theta} \mathcal{L}_{\text{BC-PMI}}(\theta) = \argmin_{\theta} \mathcal{L}_{\text{BC}}(\theta).
\end{equation}
(b). When the conditional entropy $H(Z|S,A)$ equals $0$, it indicates that there is a significant distinction in the trajectory style corresponding to all the state-action pairs. In this case, the BC-PMI objective degenerates to the behavior cloning on each style:
\begin{equation}
\label{eq_prop_b}
\mathcal{L}_{\text{BC-PMI}}(\theta) = \sum_{i=1}^K \mathcal{L}_{\mathrm{BC}}^{(i)}(\theta),
\end{equation}
where $\mathcal{L}_{\mathrm{BC}}^{(i)}(\theta)$ is the behavior cloning loss on the subset of data with style label $i$. 
\end{proposition}
\textit{Proof.} Refer to Appendix~\ref{pf1}.

Equation \ref{eq_prop_b} means when $H(Z|S,A)$ equals to $0$, the BC-PMI objection function can be viewed as separating the $(s,a)$ by the style $z$ and optimize it respectively. 
This is associated with the clustering-based behavior cloning objective, where each trajectory is first assigned to a specific style cluster based on its style label, and then behavior cloning is performed within each cluster.

The above analysis provides insights into the mechanisms of BC-PMI. In practice, the mutual information $I(Z;S,A)$ often lies between the two extremes of $0$ and $H(Z)$.
By adjusting the weight of a state-action sample in learning a style-conditioned policy, BC-PMI can smoothly interpolate between vanilla behavior cloning and clustering behavior cloning, allowing it to use the expert data more effectively than previous methods that treat each state-action sample with equal importance.

\subsection{Practical Implementation}

To practically estimate the PMI values in Eq.(\ref{eq_wwbc}), we employ the Mutual Information Neural Estimation (MINE)~\citep{belghazi2018mutual} method. MINE is a neural network-based approach that can effectively estimate mutual information between high-dimensional random variables. By leveraging the Donsker-Varadhan representation~\citep{donsker1983asymptotic} of the Kullback-Leibler (KL) divergence, MINE allows for the estimation of mutual information using neural networks.

Let $T_\phi: \mathcal{S}\times\mathcal{A}\times\mathcal{Z} \to \mathbb{R}$ be a neural network parameterized by $\phi$. The mutual information between $(s,a)$ and $z$ can be estimated as:
\begin{equation}
    \mathcal{I}(s,a;z) \geq \mathop{\mathbb{E}}\limits_{(s,a,z)\sim \mathcal{D}_e}\left[T_\phi(s,a,z)\right] - \log\left[\mathop{\mathbb{E}}\limits_{(s,a)\sim \mathcal{D}_e, \bar z\sim p(z)}\left[e^{T_\phi(s,a,\bar z)}\right]\right].
    \label{eq_mine}
\end{equation}
The neural network $T_\phi$ is trained to maximize the lower bound in Eq.(\ref{eq_mine}), which is equivalent to minimizing the KL divergence between the joint distribution $p(s,a,z)$ and the product of marginals $p(s,a)p(z)$. According to the proof of Theorem 1 in MINE~\citep{belghazi2018mutual}, the optimal solution for $T_\phi$ is:
\begin{equation}
    T_\phi^*(s,a,z) = \log\frac{p(s,a,z)}{p(s,a)p(z)} = \log\frac{p(z|s,a)}{p(z)},
\end{equation}
which is exactly the PMI we aim to estimate.

In practice, we can train the MINE network $T_\phi$ using samples from the expert demonstrations $\mathcal{D}_e$ and the style distribution $p(z)$. The training objective for $T_\phi$ is:
\begin{equation}
    \max_\phi \mathop{\mathbb{E}}\limits_{(s,a,z)\sim \mathcal{D}_e}\left[T_\phi(s,a,z)\right] - \log\left[\mathop{\mathbb{E}}\limits_{(s,a)\sim \mathcal{D}_e, \bar z\sim p(z)}\left[e^{T_\phi(s,a,\bar z)}\right]\right].
    \label{eq_mine_train}
 \end{equation}
By optimizing Eq.(\ref{eq_mine_train}), we obtain an approximation of the PMI values, which can be used as the weights in the behavioral cloning objective in Eq.(\ref{eq_wwbc}).
The weights can be denoted as:
\begin{equation}
    \sigma(s,a,z) = \exp[T_\phi^*(s,a,z)].
\end{equation}

Furthermore, in order to reduce the variance of the gradient in optimizing Eq.(\ref{eq_wwbc}), we can subtract an optimal baseline from the weight, like what has been done in A3C~\citep{mnih2016asynchronous}. This turns Eq.(\ref{eq_wwbc}) to the following objective:


\begin{equation}
    \min_\pi \mathop{\mathbb{E}}\limits_{(s,a,z)\sim \mathcal{D}_e}\Big[ -\log\pi(a|s,z)\cdot \big[\textcolor{royalblue}{\exp(T_\phi^*(s,a,z))\textcolor{amaranth}{-\widetilde{b}}}\big]  \Big],
    \label{eq_advwbc}
\end{equation}

where 
$\widetilde{b}=\mathop{\mathbb{E}}\limits_{(s,a,z)\sim \mathcal{D}_e}\Big[\exp(T_\phi^*(s,a,z))\Big]$.
In practice, we can estimate $\widetilde{b}$ using the moving average.
The pseudo-code for the BC-PMI algorithm is shown in Algorithm~\ref{alg}.

\section{Experiments}

In this experimental section, we aim to address the following questions:
\begin{enumerate}[Q1.]
    \item Can our method recover diverse and controllable policies from diverse style trajectories?
    \item Does the PMI weight have interpretability and improve policy diversity and style calibration to the algorithm?
    \item Can our method cope with complex, real-world tasks, particularly those that involve extensive datasets derived from human participants?
\end{enumerate}

\begin{algorithm}[H]
\caption{Behavioral Cloning with Pointwise Mutual Information Weighting (BC-PMI)}
\label{alg}
\begin{algorithmic}
\State Initial the parameters $\phi$ of the neural network $T$, the parameters $\theta$ of the policy $\pi$;
\State Given the expert demonstrations $\mathcal{D}_e$;
\For{$i=0,1,2,\dots$}
    \State Draw $b$ minibatch samples from the joint distribution:
    $(\{s,a\}^{(1)},z^{(1)}),\dots,(\{s,a\}^{(b)},z^{(b)})\sim 
    \mathcal{D}_e$;
    \State Draw $b$ samples from the $Z$ marginal distribution:
    $\bar z^{(1)},\dots,\bar z^{(b)}\sim P(Z)$;
    \State Update the neural network $T$ using:
    \begin{equation}
    \max_\phi \frac{1}{b}\sum_{i=1}^b T_\phi(\{s,a\}^{(i)},z^{(i)})-\log(\frac{1}{b}\sum_{i=1}^b e^{T_{\phi}(\{(s,a)\}^{(i)},\bar z^{(i)})}).
    \end{equation}
\EndFor
\For{$i=0,1,2,\dots$}
    \State Sample a random minibatch of $N$ state-action pairs from $\mathcal{D}_e$;
    \State Update the policy $\pi$ with the PMI weighting using:
    \begin{equation}
    \min_\pi \mathop{\mathbb{E}}\limits_{(s,a,z)\sim \mathcal{D}_e}\Big[ -\log\pi(a|s,z)\cdot \big[\exp(T_\phi^*(s,a,z))-\widetilde{b}\big]  \Big].
    \end{equation}
\EndFor
\end{algorithmic}
\end{algorithm}

\subsection{Styles and Baselines}

The dimension of the style needs to be specified if the style needs to be controllable.
Our method can handle both style-labeled data and data without style labels.
For the latter, getting the style labels can be expensive when relying on manual annotations and uncontrollable when using unsupervised approaches.
Instead, the programmable labeling functions~\citep{ratner2016data, zhan2020learning} can be used to automatically generate style labels.


We compare PMI-BC to the following baselines: (1) \textbf{BC}, which directly imitates expert actions across all styles; (2) \textbf{CBC}, which separates trajectories of different styles and uses BC to imitate each style separately; (3) \textbf{CTVAE}, which is the conditional version of TVAEs~\citep{wang2017robust}; (4) \textbf{InfoGAIL}~\citep{li2017infogail}, which infers the latent style of trajectories by maximizing the mutual information between the latent codes and trajectories; (5) \textbf{SORL}~\citep{mao2023stylized}\footnote{For the SORL algorithm, we only used the EM clustering part.}, which use Expectation-Maximization (EM) algorithm to classify the trajectories from the heterogeneous dataset into clusters where each represents a distinct and dominant motion style.
Specifically, we first introduce a toy example called Circle 2D to provide a simple visualization and analysis.
Then in Atari games, detailed analysis and validation were provided regarding the PMI weights.
Lastly, we evaluate all these baseline methods in Section~\ref{exp_bb} to illustrate the effectiveness of BC-PMI. 

\subsection{Circle 2D}

The Circle 2D environment is a 2D plane where an agent can freely move at a constant velocity by selecting its direction, denoted as $p_t$, at time step $t$.
For the agent, the observation at time step $t$ includes the state information from time step $t-4$ to $t$. 
The offline expert trajectories consist of four different styles, each generated by a random expert policy.
The expert policy generates trajectories that resemble circular patterns after a period of translation (75 time steps).
This design aims to introduce partial diversity in the trajectories. 
In this environment, each episode consists of 300 time steps.
If the first loop around the circle is completed before reaching 300 steps, the agent continues circling until the end of the episode.
Hence, in this scenario, there is minimal difference in the offline trajectories during the first 75 steps, and the trajectory differences vary as the agent's position progresses after 75 steps. 
During the imitation learning training process, the expert trajectories used are noisy, meaning there is randomness introduced in both the sampled actions and the environment.

\begin{figure}[htbp]
  \centering
  \subcaptionbox{Expert(without noise)\label{fig_circle_exp}}
    {\includegraphics[width=0.23\linewidth]{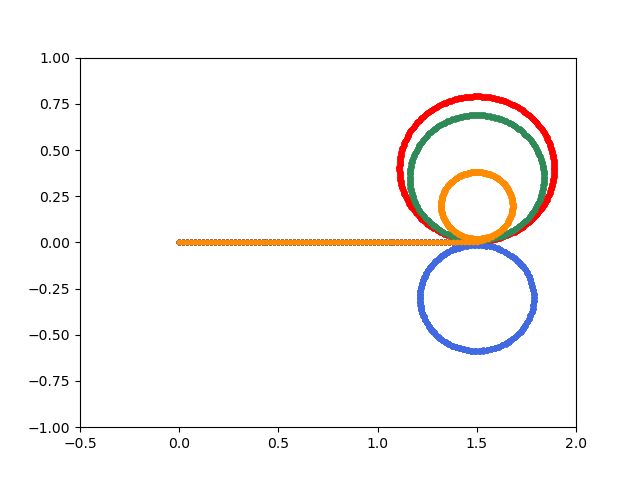}}
  \subcaptionbox{Vanilla BC\label{fig_circle_bc}}
    {\includegraphics[width=0.23\linewidth]{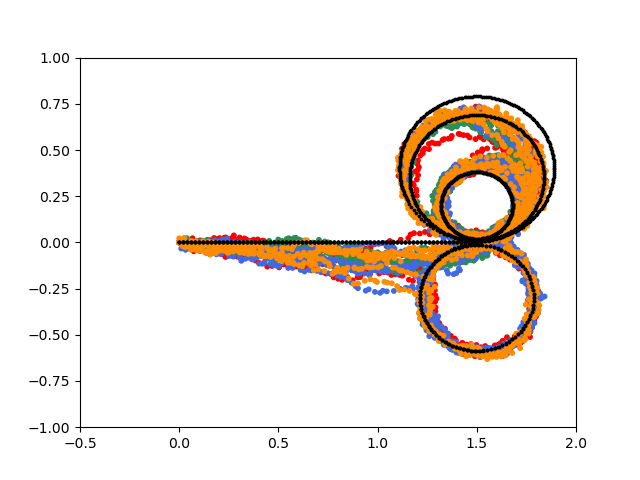}}
  \subcaptionbox{Conditional BC\label{fig_circle_cbc}}
    {\includegraphics[width=0.23\linewidth]{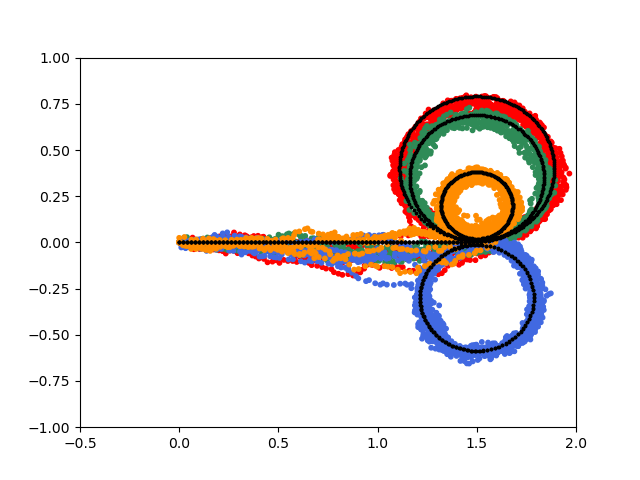}}
  \subcaptionbox{BC-PMI \label{fig_circle_rbc}}
    {\includegraphics[width=0.23\linewidth]{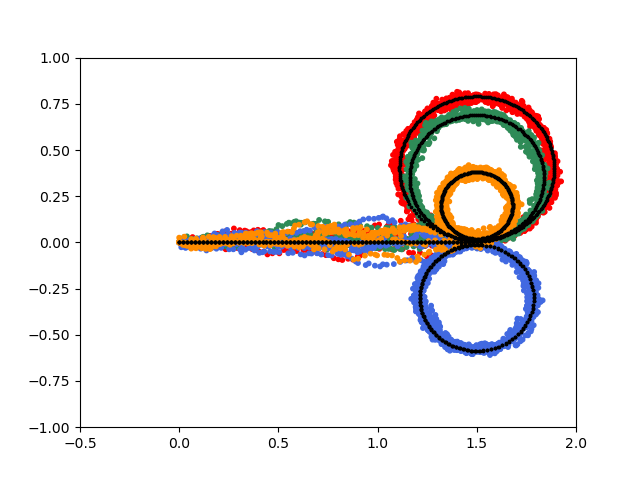}}
  \caption{Visualized comparison of trajectories generated by different policies.}
  \label{fig_circle}
\end{figure}

This task is a toy example that can help the readers intuitively understand our motivation.
In this task, the agent's behavioral diversity is highly related to the curvature of the circle moving after 75 steps and has nothing to do with the linear movement part.
Therefore, $(s,a)$ that moves according to the curvature of a specific style after 75 steps will be assigned a higher PMI, and the linear motion part or the movement curvature which are not related to the style will be assigned a lower PMI.
The visualized results are shown in Figure~\ref{fig_circle}.
By introducing PMI, the calibration of the BC policies has been further improved.
Table~\ref{table_circle} is a comparison of numerical results.
The metrics used are as follows: (1)\textbf{DTW}, which is Dynamic Time Warping, a dynamic programming algorithm for calculating the similarity of sequences of different lengths; (2)\textbf{ED}, which is Euclidean Distance; (3)\textbf{KL}, which is used to calculate the state-action distribution difference of the trajectory.

\begin{table}[htbp]
    \caption{Comparison of style calibration across different metrics in Circle 2D.}
    \label{table_circle}

  \centering
  \begin{tabular}{c|c|ccc}
    \toprule
     \textbf{Style} & \textbf{Metrics} & \textbf{BC} & \textbf{CBC} &  \textbf{BC-PMI}  \\
    \midrule\midrule
    \multirow{3}{*}{Class 1 (Red)} & DTW & 88.590 $\pm$ 9.329 & 8.130 $\pm$ 0.732 & \textbf{7.511 $\pm$ 0.545}  \\
    & ED & 133.100 $\pm$ 20.210 & 22.039 $\pm$ 1.383  & \textbf{21.591 $\pm$ 1.091} \\
    & KL & 8.037 $\pm$ 0.870 & 0.044 $\pm$ 0.006 & \textbf{0.037 $\pm$ 0.004} \\

    \midrule
    \multirow{3}{*}{Class 2 (Green)} & DTW & 45.729 $\pm$ 5.293 & 7.695 $\pm$ 0.814  & \textbf{7.631 $\pm$ 0.821} \\
    & ED & 88.180 $\pm$ 7.541 & \textbf{35.355 $\pm$ 2.691} & 35.777 $\pm$ 2.912 \\
    & KL & 1.111 $\pm$0.072 & 0.043 $\pm$ 0.003  & \textbf{0.037 $\pm$ 0.003} \\

    \midrule
    \multirow{3}{*}{Class 3 (Orange)} & DTW & 77.619 $\pm$ 7.812 & 18.972 $\pm$ 2.323  & \textbf{11.576 $\pm$ 1.788}\\
    & ED & 130.839 $\pm$ 10.322 & 107.833 $\pm$ 8.213  & \textbf{74.267 $\pm$ 8.002} \\
    & KL & 16.839 $\pm$ 1.022 & 0.271 $\pm$ 0.031 & \textbf{0.135 $\pm$ 0.014} \\

    \midrule
    \multirow{3}{*}{Class 4 (Blue)} & DTW & 69.527 $\pm$ 10.924 & 7.755 $\pm$ 1.832 & \textbf{7.577 $\pm$ 1.05}1\\
    & ED & 97.066 $\pm$ 14.239 & 27.230 $\pm$ 2.110 & \textbf{26.749 $\pm$ 2.347} \\
    & KL & 36.238 $\pm$ 2.981 & 0.401 $\pm$ 0.061  & \textbf{0.219 $\pm$ 0.019} \\

    \bottomrule
  \end{tabular}
\end{table}

\subsection{Atari Games}

In this experiment, we concentrate on three widely recognized Atari games: Alien, MsPacman and SpaceInvaders. 
The datasets utilized in this study are sourced from Atari-Head~\citep{zhang2018agil, zhang2020atari}, an extensive collection of human game-play data.
Atari-Head are meticulously recorded in a semi-frame-by-frame manner, ensuring high data quality and granularity, which facilitates in-depth analysis and robust evaluation of our proposed method.

This experiment consists of three parts.
Firstly, we demonstrate the convergence of the lower bound of mutual information in Eq.(\ref{eq_mine}), which indicates the relevance between state-action pairs and styles.
Secondly, we provide interpretability of PMI weights by assessing the extent to which they appropriately reflect the influence of the current $(s, a)$ pair on the style.
Lastly, we evaluate the controllability of the BC-PMI policy, which refers to the ability of the policy to act according to a given style once it is specified.

In the Alien and MsPacman game, we employ two styles: the area style and the range style.
The former divides the map into four distinct areas, distinguishing the agent's preferences for moving toward each area.
The latter models the agent's displacement trajectory on the map as a Gaussian distribution, differentiating the variance of the distribution.
A higher variance indicates a tendency for the agent to move across areas, while a lower variance indicates a preference for movement within a single area. 
In the SpaceInvader game, we also utilize two styles: the firing rate style and the area style.
For further details, please refer to Appendix~\ref{style_sp}.

\begin{wrapfigure}[12]{r}{7cm}
\centering
\setlength{\abovecaptionskip}{0.cm}
\vspace*{-5mm}
\includegraphics[width=0.5\textwidth]{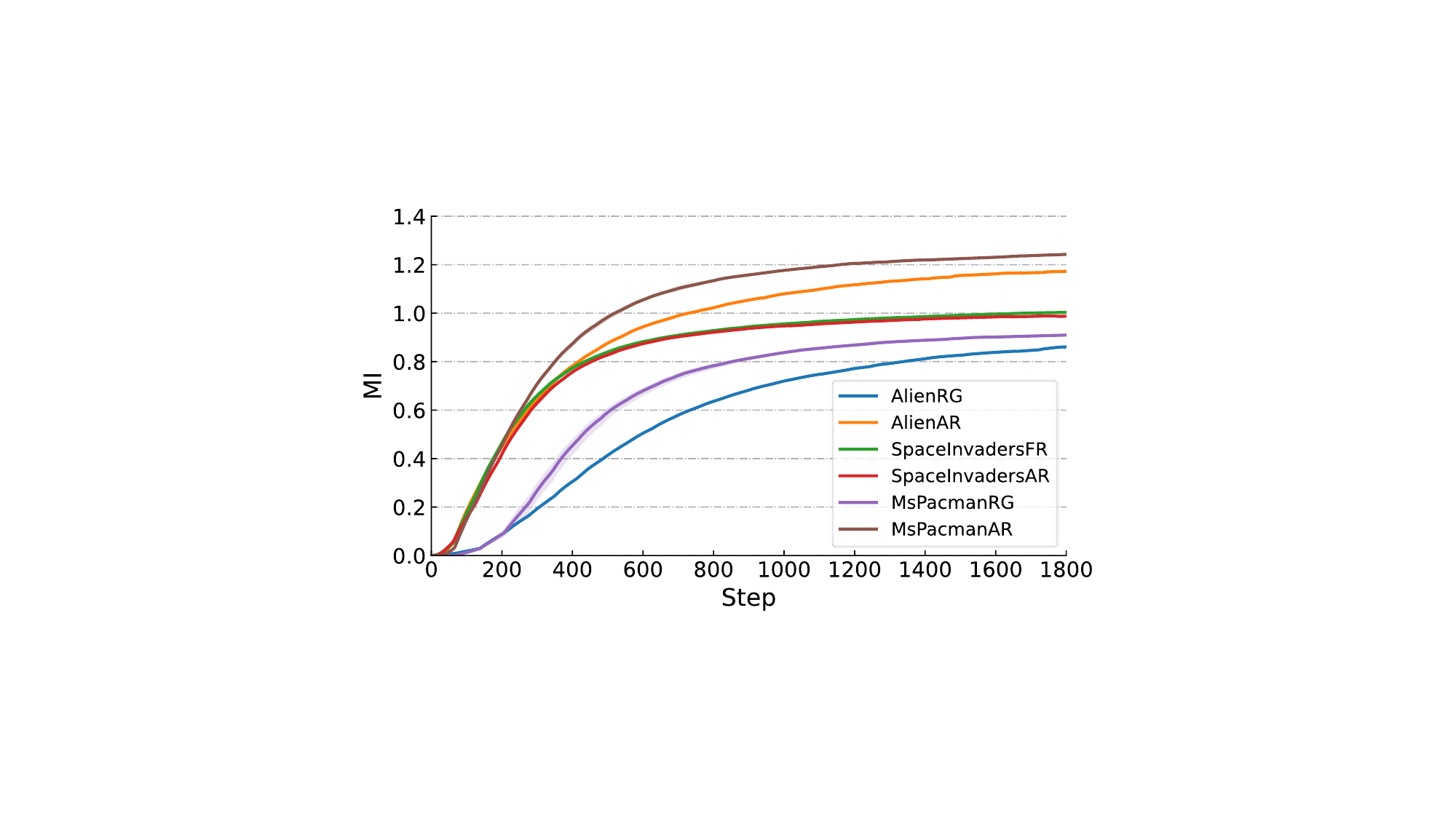}
\caption{MI between state-action pairs and styles. FR: Fire Rate style; AR: Movement Area style; RG: Movement Range style.}
\label{mi_loss}
\end{wrapfigure}
The results in Figure~\ref{mi_loss} indicate a high relevance between the agent's state-action distribution and the style.
Moreover, as the relevance increases, the converged MI values also increase. 
The MsPacmanAR style divides the map into four areas: top-left, top-right, bottom-left, and bottom-right.
The agent's trajectory exhibits clear distinctions under this style, which is reflected in the larger converged MI values, while the MsPacmanRG style represents the style based on the range of movement, which is less distinct compared to the movement area style.


\begin{figure}[htbp]
  \centering
  \includegraphics[width=\linewidth]{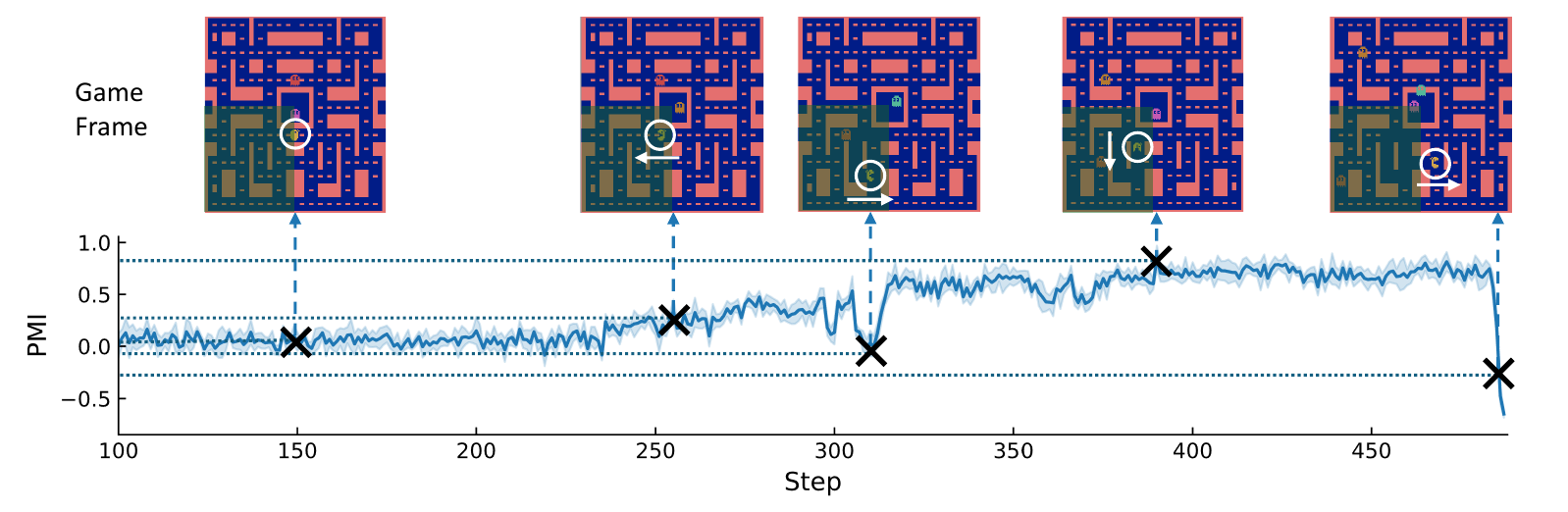}
  \caption{The PMI weight values related to style for each frame along a trajectory, with the corresponding style being the bottom-left area style (green color) in the game frame. The agent in the game frame is indicated by a white circle, and the corresponding actions are indicated by white arrows. No arrow indicates that the action is NOOP.}
  \label{pmis}
\end{figure}

As shown in Figure~\ref{pmis}, in the first frame, the agent is located at the edge of the area and does not take any action, resulting in a near-zero relevance with the style. In the second and fourth frames, the agent moves to the corresponding area and exhibits a tendency to continue moving towards that area, leading to a higher relevance with the style and higher PMI values. Conversely, in the third and fifth frames, the agent is positioned at the edge or outside of the style area and shows a tendency to move away from the style area, resulting in a lower relevance with the style.

To verify whether different styles of policies affect the cumulative reward, we conducted experiments in three different Atari environments, as shown in Table~\ref{table_atari_reward}.
The results indicate that most styles, such as those related to position and range of movement, do not impact the cumulative reward of the agent.
However, some styles that are highly correlated with scoring can significantly influence the cumulative reward.
For instance, in Space Invaders, the Fire style, characterized by a higher firing frequency, significantly increases the agent's cumulative reward, while a lower firing frequency reduces the cumulative reward.

\begin{table}[htbp]
    \caption{Comparison of the cumulative reward across different styles in Atari. BC-PMI-x means the reward of the x-th style.}
    \label{table_atari_reward}
    
  \centering
  
  \begin{tabular}{p{1.2cm}|p{0.8cm}|ccccc}
    \toprule
    \textbf{Env} & \textbf{Style} & \textbf{\quad BC\quad } & \textbf{ BC-PMI-1 } &  \textbf{BC-PMI-2} & \textbf{BC-PMI-3} & \textbf{BC-PMI-4} \\
    \midrule\midrule
    \multirow{2}{*}{Alien} & Pos & 460.5$\pm$173.4 & 416.0$\pm$160.8 & 432.0$\pm$174.7 & 500.3$\pm$211.7 & 496.1$\pm$191.5 \\
    & Range & 460.5$\pm$173.4 & 440.0$\pm$217.0 & 501.2$\pm$247.9 & 423.2$\pm$240.9 & - \\
    \midrule
    Ms & Pos & 701.5$\pm$182.4 & 665.4$\pm$216.3 & 803.5$\pm$211.2 & 674.3$\pm$146.5 & 666.2$\pm$172.2 \\
    Pacman & Range & 701.5$\pm$182.4 & 636.1$\pm$265.1 & 759.5$\pm$280.6 & 738.3$\pm$260.8 & - \\
    \midrule
    Space & Pos & 208.2$\pm$73.0 & 178.9$\pm$54.5 & 187.0$\pm$54.9 & 200.2$\pm$56.4 & - \\
    Invaders & Fire & 208.2$\pm$73.0 & 94.8$\pm$47.2 & 207.9$\pm$70.3 & 270.6$\pm$86.7 & - \\
    \bottomrule
  \end{tabular}
\end{table}

Finally, we compared the calibration of several different style policies across these Atari environments, as shown in Table~\ref{table_atari_calibration}.
In styles related to position and firing, BC-PMI outperformed all baselines.
In styles related to range, BC-PMI achieved performance comparable to CTVAE.
The reason is that the mutual information distinguishing capability of styles related to position and firing is higher (refer to Figure~\ref{mi_loss}).
Consequently, the trained MINE network can assign more accurate PMI values to state-action pairs, resulting in higher accuracy for the corresponding styles.

\begin{table}[htbp]
    \caption{Comparison of style calibration across different styles in Atari}
    \label{table_atari_calibration}
    
  \centering
  \begin{tabular}{p{1.2cm}|p{0.8cm}|cccccc}
    \toprule
    \textbf{Env} & \textbf{Style} & \textbf{\quad BC\quad } & \textbf{ CBC } &  \textbf{CTVAE} & \textbf{InfoGAIL} & \textbf{SORL} & \textbf{BC-PMI} \\
    \midrule\midrule
    \multirow{2}{*}{Alien} & Pos & 0.23$\pm$0.11 & 0.43$\pm$0.04 & 0.52$\pm$0.04 & 0.34$\pm$0.11 & 0.31$\pm$0.07 & \textbf{0.61$\pm$0.08} \\
    & Range & 0.31$\pm$0.09 & 0.51$\pm$0.13 & \textbf{0.57$\pm$0.09} & 0.31$\pm$0.11 & 0.27$\pm$0.07 & 0.55$\pm$0.12 \\
    \midrule
    MS & Pos & 0.17$\pm$0.05 & 0.48$\pm$0.07 & 0.55$\pm$0.04 & 0.33$\pm$0.05 & 0.27$\pm$0.08 & \textbf{0.57$\pm$0.07} \\
    Pacman & Range & 0.33$\pm$0.09 & 0.53$\pm$0.11 & \textbf{0.61$\pm$0.05} & 0.29$\pm$0.09 & 0.30$\pm$0.12 & 0.56$\pm$0.14 \\
    \midrule
    Space & Pos & 0.36$\pm$0.06 & 0.71$\pm$0.08 & 0.66$\pm$0.10 & 0.41$\pm$0.06 & 0.37$\pm$0.04 & \textbf{0.83$\pm$0.06} \\
    Invaders & Fire & 0.27$\pm$0.07 & 0.41$\pm$0.08 & 0.47$\pm$0.06 & 0.36$\pm$0.08 & 0.34$\pm$0.09 & \textbf{0.51$\pm$0.10} \\
    \bottomrule
  \end{tabular}
\end{table}



\subsection{Professional Basketball Player Dataset}
\label{exp_bb}
In this experiment, we validate our method on the dataset of a collection of professional basketball player trajectories~\citep{zhan2020learning} with the goal of recovering policies that can generate trajectories with diverse player-movement styles.
The basketball trajectories are collected from tracking real players in the NBA.
We primarily focus on two movement styles: (1) The \textbf{Destination}, which is the distance from the final position to a fixed destination on the
court (e.g. the basket), and (2) The \textbf{Curvature}, which measures
the player’s propensity to change directions.

\begin{figure}[htbp]
  \centering
  \includegraphics[width=.9\linewidth]{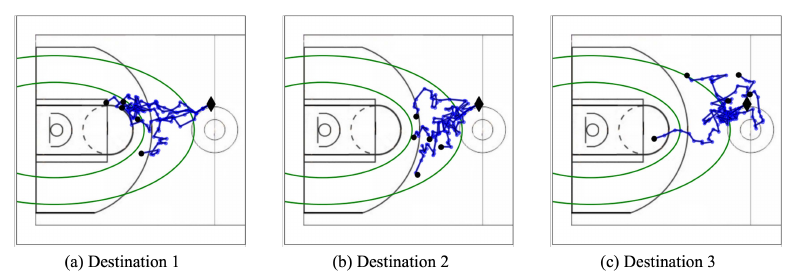}
  \caption{Visualization of different destination styles.}
  \label{bb_dest}
\end{figure}

\begin{figure}[htbp]
  \centering
  \includegraphics[width=.9\linewidth]{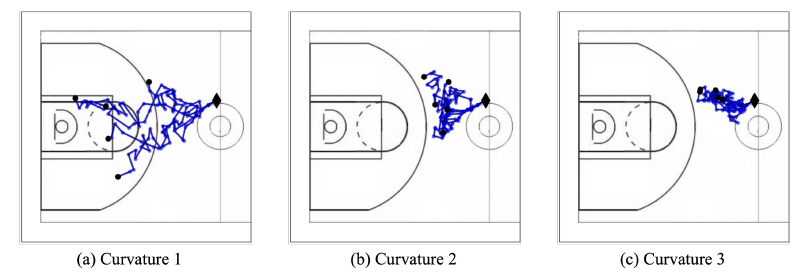}
  \caption{Visualization of different curvature styles.}
  \label{bb_cur}
\end{figure}

Figure~\ref{bb_dest} and Figure~\ref{bb_cur} present the calibration results of different styles for the BC-PMI algorithm in the dimensions of \textit{Destination} and \textit{Curvature}.
The environment is a half-court basketball setting, where each player's trajectory can be categorized into different styles based on their movement destination and curvature.
In Figure~\ref{bb_dest}, Destination 1 is close to the ball frame, Destination 2 is in the middle count, and Destination 3 is far from the ball frame.
The three destinations are separated by green lines in the figure.
Similarly, three different movement curvature styles are illustrated in Figure~\ref{bb_cur}.
The results indicate that the BC-PMI algorithm can effectively imitate policies of different movement styles from real human data.

\begin{table}[htbp]
    \caption{Comparison of style calibration (\%) across different styles in Basketball}
    \label{table_bb}
    
  \centering
  \begin{tabular}{c|c|cccccc}
    \toprule
    \textbf{Style} & \textbf{Class} & \textbf{\quad BC\quad } & \textbf{ CBC } &  \textbf{CTVAE} & \textbf{InfoGAIL} & \textbf{SORL} & \textbf{BC-PMI}  \\
    \midrule\midrule
    \multirow{3}{*}{Destination} & Class 1 & 31.1 & 81.6 & 78.3 & 79.6 & 80.9 & \textbf{93.3} \\
    & Class 2 & 42.7 & 84.4 & 81.2 & 75.5 & 76.1 & \textbf{92.7} \\
    & Class 3 & 19.4 & 73.1 & 77.8 & 74.5 & 79.5 & \textbf{89.8} \\
    \midrule
    \multirow{3}{*}{Curvature} & Class 1 & 17.1 & 67.3 & 58.4 & 60.9 & 51.6 & \textbf{79.4} \\
    & Class 2 & 30.6 & 66.2 & 58.7 & 57.2 & 49.4 & \textbf{80.9} \\
    & Class 3 & 41.8 & 71.6 & 62.1 & 62.7 & 53.8 & \textbf{81.6} \\
    \bottomrule
  \end{tabular}
\end{table}

We compare style calibration for 3 classes of \textit{Destination} and 3 classes of \textit{Curvature} in Table~\ref{table_bb}.
Due to the lack of style label information, BC can only learn an approximately average strategy, so it should be evenly distributed among various styles.
However, due to factors such as initialization position, the samples in each category do not strictly follow a uniform distribution.
Among the various methods with diversity mechanisms, the BC-PMI method, which focuses on state-action pairs highly relevant to the style, surpasses other baselines in terms of style calibration.

\section{Discussion}

\subsection{Conclusion}
In this paper, we investigate how to recover diverse policies from a set of expert trajectories.
We propose a new diverse policy recovering method by leveraging the relevance of the state-action pair with the trajectory styles.
The highlight of our method lies in our approach to the problem of policy diversity from a different perspective, which involves the introduction of Pointwise Mutual Information to model the relevance between each $(s, a)$ pair and the style.
By utilizing a unique and straightforward approach, we achieved results that surpassed previous state-of-the-art methods.

\subsection{Limitations and Future Work}
In our setting, our goal is to recover policies from diverse offline data, assuming that the data within the trajectories already meet the performance requirements.
If the policy that generates the offline trajectories performs poorly, our method cannot further enhance the performance of the learned policy.
Hence, we consider the combination of PMI and RL as future research, in which we believe the PMI weight can be associated with the offline RL method to promote diversity while improving performance. 

\bibliography{iclr2025_conference}

\begin{thebibliography}{38}
\providecommand{\natexlab}[1]{#1}
\providecommand{\url}[1]{\texttt{#1}}
\expandafter\ifx\csname urlstyle\endcsname\relax
  \providecommand{\doi}[1]{doi: #1}\else
  \providecommand{\doi}{doi: \begingroup \urlstyle{rm}\Url}\fi

\bibitem[Achiam et~al.(2018)Achiam, Edwards, Amodei, and Abbeel]{achiam2018variational}
Joshua Achiam, Harrison Edwards, Dario Amodei, and Pieter Abbeel.
\newblock Variational option discovery algorithms.
\newblock \emph{arXiv preprint arXiv:1807.10299}, 2018.

\bibitem[Arora \& Doshi(2021)Arora and Doshi]{arora2021survey}
Saurabh Arora and Prashant Doshi.
\newblock A survey of inverse reinforcement learning: Challenges, methods and progress.
\newblock \emph{Artificial Intelligence}, 297:\penalty0 103500, 2021.

\bibitem[Belghazi et~al.(2018)Belghazi, Baratin, Rajeshwar, Ozair, Bengio, Courville, and Hjelm]{belghazi2018mutual}
Mohamed~Ishmael Belghazi, Aristide Baratin, Sai Rajeshwar, Sherjil Ozair, Yoshua Bengio, Aaron Courville, and Devon Hjelm.
\newblock Mutual information neural estimation.
\newblock In \emph{International Conference on Machine Learning}, pp.\  531--540. PMLR, 2018.

\bibitem[Bouma(2009)]{bouma2009normalized}
Gerlof Bouma.
\newblock Normalized (pointwise) mutual information in collocation extraction.
\newblock \emph{Proceedings of GSCL}, 30:\penalty0 31--40, 2009.

\bibitem[Br{\'e}maud(2012)]{bremaud2012introduction}
Pierre Br{\'e}maud.
\newblock \emph{An introduction to probabilistic modeling}.
\newblock Springer Science \& Business Media, 2012.

\bibitem[Campos et~al.(2020)Campos, Trott, Xiong, Socher, Gir{\'o}-i Nieto, and Torres]{campos2020explore}
V{\'\i}ctor Campos, Alexander Trott, Caiming Xiong, Richard Socher, Xavier Gir{\'o}-i Nieto, and Jordi Torres.
\newblock Explore, discover and learn: Unsupervised discovery of state-covering skills.
\newblock In \emph{International Conference on Machine Learning}, pp.\  1317--1327. PMLR, 2020.

\bibitem[Chen et~al.(2016)Chen, Duan, Houthooft, Schulman, Sutskever, and Abbeel]{chen2016infogan}
Xi~Chen, Yan Duan, Rein Houthooft, John Schulman, Ilya Sutskever, and Pieter Abbeel.
\newblock Infogan: Interpretable representation learning by information maximizing generative adversarial nets.
\newblock \emph{Advances in Neural Information Processing Systems}, 29, 2016.

\bibitem[Church \& Hanks(1990)Church and Hanks]{church1990word}
Kenneth Church and Patrick Hanks.
\newblock Word association norms, mutual information, and lexicography.
\newblock \emph{Computational Linguistics}, 16\penalty0 (1):\penalty0 22--29, 1990.

\bibitem[Codevilla et~al.(2018)Codevilla, M{\"u}ller, L{\'o}pez, Koltun, and Dosovitskiy]{codevilla2018end}
Felipe Codevilla, Matthias M{\"u}ller, Antonio L{\'o}pez, Vladlen Koltun, and Alexey Dosovitskiy.
\newblock End-to-end driving via conditional imitation learning.
\newblock In \emph{2018 IEEE international conference on robotics and automation (ICRA)}, pp.\  4693--4700. IEEE, 2018.

\bibitem[Donsker \& Varadhan(1983)Donsker and Varadhan]{donsker1983asymptotic}
Monroe~D Donsker and SR~Srinivasa Varadhan.
\newblock Asymptotic evaluation of certain markov process expectations for large time. iv.
\newblock \emph{Communications on Pure and Applied Mathematics}, 36\penalty0 (2):\penalty0 183--212, 1983.

\bibitem[Eysenbach et~al.(2018)Eysenbach, Gupta, Ibarz, and Levine]{eysenbach2018diversity}
Benjamin Eysenbach, Abhishek Gupta, Julian Ibarz, and Sergey Levine.
\newblock Diversity is all you need: Learning skills without a reward function.
\newblock \emph{arXiv preprint arXiv:1802.06070}, 2018.

\bibitem[Fu et~al.(2017)Fu, Luo, and Levine]{fu2017learning}
Justin Fu, Katie Luo, and Sergey Levine.
\newblock Learning robust rewards with adversarial inverse reinforcement learning.
\newblock \emph{arXiv preprint arXiv:1710.11248}, 2017.

\bibitem[Fujimoto et~al.(2019)Fujimoto, Meger, and Precup]{fujimoto2019off}
Scott Fujimoto, David Meger, and Doina Precup.
\newblock Off-policy deep reinforcement learning without exploration.
\newblock In \emph{International conference on machine learning}, pp.\  2052--2062. PMLR, 2019.

\bibitem[Hausman et~al.(2017)Hausman, Chebotar, Schaal, Sukhatme, and Lim]{hausman2017multi}
Karol Hausman, Yevgen Chebotar, Stefan Schaal, Gaurav Sukhatme, and Joseph~J Lim.
\newblock Multi-modal imitation learning from unstructured demonstrations using generative adversarial nets.
\newblock \emph{Advances in neural information processing systems}, 30, 2017.

\bibitem[Hjelm et~al.(2018)Hjelm, Fedorov, Lavoie-Marchildon, Grewal, Bachman, Trischler, and Bengio]{hjelm2018learning}
R~Devon Hjelm, Alex Fedorov, Samuel Lavoie-Marchildon, Karan Grewal, Phil Bachman, Adam Trischler, and Yoshua Bengio.
\newblock Learning deep representations by mutual information estimation and maximization.
\newblock \emph{arXiv preprint arXiv:1808.06670}, 2018.

\bibitem[Ho \& Ermon(2016)Ho and Ermon]{ho2016generative}
Jonathan Ho and Stefano Ermon.
\newblock Generative adversarial imitation learning.
\newblock \emph{Advances in Neural Information Processing Systems}, 29, 2016.

\bibitem[Hussein et~al.(2017)Hussein, Gaber, Elyan, and Jayne]{hussein2017imitation}
Ahmed Hussein, Mohamed~Medhat Gaber, Eyad Elyan, and Chrisina Jayne.
\newblock Imitation learning: A survey of learning methods.
\newblock \emph{ACM Computing Surveys (CSUR)}, 50\penalty0 (2):\penalty0 1--35, 2017.

\bibitem[Kostrikov et~al.(2021)Kostrikov, Nair, and Levine]{kostrikov2021offline}
Ilya Kostrikov, Ashvin Nair, and Sergey Levine.
\newblock Offline reinforcement learning with implicit q-learning.
\newblock \emph{arXiv preprint arXiv:2110.06169}, 2021.

\bibitem[Li et~al.(2017)Li, Song, and Ermon]{li2017infogail}
Yunzhu Li, Jiaming Song, and Stefano Ermon.
\newblock Infogail: Interpretable imitation learning from visual demonstrations.
\newblock \emph{Advances in Neural Information Processing Systems}, 30, 2017.

\bibitem[Manning \& Schutze(1999)Manning and Schutze]{manning1999foundations}
Christopher Manning and Hinrich Schutze.
\newblock \emph{Foundations of statistical natural language processing}.
\newblock MIT press, 1999.

\bibitem[Mao et~al.(2023)Mao, Wu, Chen, Hu, Jiang, Zhou, Lv, Fan, Hu, Wu, et~al.]{mao2023stylized}
Yihuan Mao, Chengjie Wu, Xi~Chen, Hao Hu, Ji~Jiang, Tianze Zhou, Tangjie Lv, Changjie Fan, Zhipeng Hu, Yi~Wu, et~al.
\newblock Stylized offline reinforcement learning: Extracting diverse high-quality behaviors from heterogeneous datasets.
\newblock In \emph{International Conference on Learning Representations}, 2023.

\bibitem[Mnih et~al.(2016)Mnih, Badia, Mirza, Graves, Lillicrap, Harley, Silver, and Kavukcuoglu]{mnih2016asynchronous}
Volodymyr Mnih, Adria~Puigdomenech Badia, Mehdi Mirza, Alex Graves, Timothy Lillicrap, Tim Harley, David Silver, and Koray Kavukcuoglu.
\newblock Asynchronous methods for deep reinforcement learning.
\newblock In \emph{International Conference on Machine Learning}, pp.\  1928--1937. PMLR, 2016.

\bibitem[Ng et~al.(2000)Ng, Russell, et~al.]{ng2000algorithms}
Andrew~Y Ng, Stuart Russell, et~al.
\newblock Algorithms for inverse reinforcement learning.
\newblock In \emph{International Conference on Machine Learning}, volume~1, pp.\ ~2, 2000.

\bibitem[Osa et~al.(2018)Osa, Pajarinen, Neumann, Bagnell, Abbeel, Peters, et~al.]{osa2018algorithmic}
Takayuki Osa, Joni Pajarinen, Gerhard Neumann, J~Andrew Bagnell, Pieter Abbeel, Jan Peters, et~al.
\newblock An algorithmic perspective on imitation learning.
\newblock \emph{Foundations and Trends{\textregistered} in Robotics}, 7\penalty0 (1-2):\penalty0 1--179, 2018.

\bibitem[Peng et~al.(2019)Peng, Kumar, Zhang, and Levine]{peng2019advantage}
Xue~Bin Peng, Aviral Kumar, Grace Zhang, and Sergey Levine.
\newblock Advantage-weighted regression: Simple and scalable off-policy reinforcement learning.
\newblock \emph{arXiv preprint arXiv:1910.00177}, 2019.

\bibitem[Pomerleau(1991)]{pomerleau1991efficient}
Dean~A Pomerleau.
\newblock Efficient training of artificial neural networks for autonomous navigation.
\newblock \emph{Neural Computation}, 3\penalty0 (1):\penalty0 88--97, 1991.

\bibitem[Ratner et~al.(2016)Ratner, De~Sa, Wu, Selsam, and R{\'e}]{ratner2016data}
Alexander~J Ratner, Christopher~M De~Sa, Sen Wu, Daniel Selsam, and Christopher R{\'e}.
\newblock Data programming: Creating large training sets, quickly.
\newblock \emph{Advances in Neural Information Processing Systems}, 29, 2016.

\bibitem[Ross et~al.(2011)Ross, Gordon, and Bagnell]{ross2011reduction}
St{\'e}phane Ross, Geoffrey Gordon, and Drew Bagnell.
\newblock A reduction of imitation learning and structured prediction to no-regret online learning.
\newblock In \emph{Proceedings of the Fourteenth International Conference on Artificial Intelligence and Statistics}, pp.\  627--635. JMLR Workshop and Conference Proceedings, 2011.

\bibitem[Shafiullah et~al.(2022)Shafiullah, Cui, Altanzaya, and Pinto]{shafiullah2022behavior}
Nur~Muhammad Shafiullah, Zichen Cui, Ariuntuya~Arty Altanzaya, and Lerrel Pinto.
\newblock Behavior transformers: Cloning $ k $ modes with one stone.
\newblock \emph{Advances in neural information processing systems}, 35:\penalty0 22955--22968, 2022.

\bibitem[Sharma et~al.(2019)Sharma, Gu, Levine, Kumar, and Hausman]{sharma2019dynamics}
Archit Sharma, Shixiang Gu, Sergey Levine, Vikash Kumar, and Karol Hausman.
\newblock Dynamics-aware unsupervised discovery of skills.
\newblock \emph{arXiv preprint arXiv:1907.01657}, 2019.

\bibitem[Sutton \& Barto(2018)Sutton and Barto]{sutton2018reinforcement}
Richard~S Sutton and Andrew~G Barto.
\newblock \emph{Reinforcement learning: An introduction}.
\newblock MIT press, 2018.

\bibitem[Wang et~al.(2017)Wang, Merel, Reed, de~Freitas, Wayne, and Heess]{wang2017robust}
Ziyu Wang, Josh~S Merel, Scott~E Reed, Nando de~Freitas, Gregory Wayne, and Nicolas Heess.
\newblock Robust imitation of diverse behaviors.
\newblock \emph{Advances in Neural Information Processing Systems}, 30, 2017.

\bibitem[Wu et~al.(2019)Wu, Tucker, and Nachum]{wu2019behavior}
Yifan Wu, George Tucker, and Ofir Nachum.
\newblock Behavior regularized offline reinforcement learning.
\newblock \emph{arXiv preprint arXiv:1911.11361}, 2019.

\bibitem[Yannakakis \& Togelius(2018)Yannakakis and Togelius]{yannakakis2018artificial}
Georgios~N Yannakakis and Julian Togelius.
\newblock \emph{Artificial intelligence and games}, volume~2.
\newblock Springer, 2018.

\bibitem[Zhan et~al.(2020)Zhan, Tseng, Yue, Swaminathan, and Hausknecht]{zhan2020learning}
Eric Zhan, Albert Tseng, Yisong Yue, Adith Swaminathan, and Matthew Hausknecht.
\newblock Learning calibratable policies using programmatic style-consistency.
\newblock In \emph{International Conference on Machine Learning}, pp.\  11001--11011. PMLR, 2020.

\bibitem[Zhang et~al.(2018)Zhang, Liu, Zhang, Whritner, Muller, Hayhoe, and Ballard]{zhang2018agil}
Ruohan Zhang, Zhuode Liu, Luxin Zhang, Jake~A Whritner, Karl~S Muller, Mary~M Hayhoe, and Dana~H Ballard.
\newblock Agil: Learning attention from human for visuomotor tasks.
\newblock In \emph{Proceedings of the European Conference on Computer Vision}, pp.\  663--679, 2018.

\bibitem[Zhang et~al.(2020)Zhang, Walshe, Liu, Guan, Muller, Whritner, Zhang, Hayhoe, and Ballard]{zhang2020atari}
Ruohan Zhang, Calen Walshe, Zhuode Liu, Lin Guan, Karl Muller, Jake Whritner, Luxin Zhang, Mary Hayhoe, and Dana Ballard.
\newblock Atari-head: Atari human eye-tracking and demonstration dataset.
\newblock In \emph{Proceedings of the AAAI conference on Artificial Intelligence}, volume~34, pp.\  6811--6820, 2020.

\bibitem[Zhu et~al.(2018)Zhu, Wang, Merel, Rusu, Erez, Cabi, Tunyasuvunakool, Kram{\'a}r, Hadsell, de~Freitas, et~al.]{zhu2018reinforcement}
Yuke Zhu, Ziyu Wang, Josh Merel, Andrei Rusu, Tom Erez, Serkan Cabi, Saran Tunyasuvunakool, J{\'a}nos Kram{\'a}r, Raia Hadsell, Nando de~Freitas, et~al.
\newblock Reinforcement and imitation learning for diverse visuomotor skills.
\newblock \emph{arXiv preprint arXiv:1802.09564}, 2018.

\end{thebibliography}
\bibliographystyle{iclr2025_conference}

\newpage
\appendix
\section{Theoretical Derivation}

\subsection{Proof of Proposition 1}
\label{pf1}

\begin{proof}
(a) We first introduce the Gibbs' inequality, which was proved in \citep{bremaud2012introduction}.
\begin{lemma}[Gibbs' inequality]
Let $p(x)$, $q(x)$ be the probability mass function, fixed $p(x)$, that the object function $-\mathbb{E}_{x\sim p(x)}\log q(x)=\sum_{x} -p(x) \log q(x) $ takes the minimum when $q(x)=p(x)$.
\end{lemma}
By Lemma 1, we know
\begin{equation}
\mathcal{L}_{BC}(\theta)=\mathop{\mathbb{E}}\limits_{(s,a)\sim \mathcal{D}_e}[-\log\pi_{\theta}(a|s)] =\mathop{\mathbb{E}}\limits_{s\sim \mathcal{D}_e} \left [\mathop{\mathbb{E}}\limits_{a|s \sim p(a|s)} [-\log\pi_{\theta}(a|s)]\right ],
\end{equation}
which takes the minimum when $\pi_{\theta}(a|s)=p(a|s)$.

When $I(Z;S,A)=0$, it means that the latent variable $Z$ is independent of the state-action pair $(S,A)$, and the PMI term in the BC-PMI objective becomes:
\begin{equation}
    \frac{p(z|s,a)}{p(z)} = \frac{p(z)}{p(z)} = 1.
\end{equation}
Hence, 
\begin{align}
    \mathcal{L}_{\mathrm{BC-PMI}}(\theta) &=\mathop{\mathbb{E}}\limits_{(s,a,z)\sim\mathcal{D}_e}[-\log\pi_{\theta}(a|s,z)] \nonumber \\
    &=\mathop{\mathbb{E}}\limits_{s\sim \mathcal{D}_e} \left[\sum_a \sum_z -p(a|s) p(z|a,s)\log\pi_{\theta}(a|s,z) \right] \nonumber \\
    &= \mathop{\mathbb{E}}\limits_{s\sim \mathcal{D}_e} \left[ \sum_z \sum_a -p(a|s) p(z) \log\pi_{\theta}(a|s,z) \right] \nonumber \\ 
    &= \mathop{\mathbb{E}}\limits_{s\sim \mathcal{D}_e} \left[ \sum_z \left[p(z) (\sum_a -p(a|s) \log\pi_{\theta}(a|s,z))\right] \right],
\end{align}
where the third equation is because $z$ is independent with $(s,a)$.
Since $p(z)$ is non-negative, when $\sum_a -p(a|s) \log\pi_{\theta}(a|s,z)$ takes the minimum for all $z$, the objective reach its the minimum. Using Lemma 1 again, we can see when $\pi_{\theta}(a|s,z)=p(a|s)$, $\mathcal{L}_{\mathrm{BC-PMI}}(\theta)$ reach its minimum. 

(b) when $H(Z|S,A)=0$, it means that given a state-action pair $(s,a)$, the style variable $Z$ can be determined with high certainty.
In other words, for any $(s,a,z)\sim\mathcal{D}_e$, we have:
\begin{equation}
p(z|s,a)=\mathbbm{1}(z=z_{s,a}),
\end{equation}
where $z_{s,a}$ is the unique style label corresponding to $(s,a)$.

Assuming there are $K$ styles in the Dataset. Let $z_\tau$ be the unique style label corresponding to trajectory $\tau$, $D_e^{(i)}$ be the subset of state-action pairs with style label $i$ and $\tau \in D_e^{(i)}$ means the whole trajectory from $D_e^{(i)}$. Denote 
\begin{equation}
    \mathcal{L}_{\mathrm{BC}}^{(i)}(\theta)= \mathop{\mathbb{E}}\limits_{(s',a')\sim D_e^{(i)}} [-\log\pi_\theta(a'|s',i)] =\frac{1}{|D_e^{(i)}|} \sum_{\tau\in D_e^{(i)}} \sum_{(s',a')\in \tau} -\log\pi_\theta(a'|s',i).
\end{equation}
We have 
\begin{align}
\mathcal{L}_{\mathrm{BC-PMI}}(\theta) 
&= \mathop{\mathbb{E}}\limits_{(s,a,z)\sim \mathcal{D}_e}
\left[ -\log\pi_\theta(a|s,z)\cdot \frac{p(z|s,a)}{p(z)} \right] 
\nonumber\\
&= \mathop{\mathbb{E}}\limits_{(s,a,z)\sim D_e}[-\log\pi_{\theta}(a|s,z)\frac{\mathbbm{1}(z=z_{s,a})}{p(z)}] \nonumber\\
& = \frac{1}{|D_e|} \sum_{\tau\in D_e} \sum_{i=1}^K \mathbbm{1}(z_{\tau}=i) \sum_{(s',a')\in \tau} [-\log\pi_\theta(a'|s',i)\cdot \frac{1}{p(z=i)}] \nonumber\\   
&= \frac{1}{|D_e|} \sum_{i=1}^K \sum_{\tau\in D_e} \sum_{(s',a')\in \tau} \mathbbm{1}(z_{\tau}=i) [-\log\pi_\theta(a'|s',i)\cdot \frac{|D_e|}{|D_e^{(i)}|}] \nonumber \\
&=\sum_{i=1}^K \sum_{\tau\in D_e^{(i)}} \frac{1}{|D_e^{(i)}|} \sum_{(s',a')\in \tau} -\log\pi_\theta(a'|s',i) \nonumber\\
&=\sum_{i=1}^K \mathcal{L}_{\mathrm{BC}}^{(i)}(\theta)
\end{align}

Note that by the definition, $\mathcal{L}_{\mathrm{BC}}^{(i)}(\theta)$ is the behavior cloning loss on the subset of data with style label $i$. Hence, the last equation above means when $H(Z|A,S)$ equals $0$, the BC-PMI objection function can be viewed as separating the $(s,a)$ by the style $z$ and optimizing it respectively. This is equivalent to the clustering-based behavior cloning objective, where each trajectory is first assigned to a specific style cluster based on its style label, and then behavior cloning is performed within each cluster.
This result reveals the close connection between BC-PMI and clustering-based behavior cloning in the extreme case of zero style overlap.
\end{proof}

\section{Implementation Details}
\subsection{Computational Resource}
All experiments in this paper are implemented with PyTorch and executed on NVIDIA Tesla T4 GPUs.
All the runs in experiments use 5 random seeds.

\subsection{Common Hyperparameters}
\begin{table}[htbp]
  \caption{Common hyperparameters setting.}
  \label{detail}
  \centering
  \begin{tabular}{ccccc}
    \toprule
    Hyperparameter& Circle 2D  & MsPacman & SpaceInvaders & Basketball \\
    \midrule
    learning rate & 0.001  & 0.001  & 0.001 & 0.0002    \\
    optimizer     & Adam    & Adam    & Adam   & Adam \\
    epoch         & 10      & 30      & 30     & 30 \\
    batch size    & 128    & 512    & 512   & 128 \\
    hidden dim    & 32    & 64    & 64   &  128 \\
    \bottomrule
  \end{tabular}
\label{table_com_para}
\end{table}

\subsection{MINE Network Structure}
\begin{figure}[htbp]
\centering
\includegraphics[width=.7\textwidth]{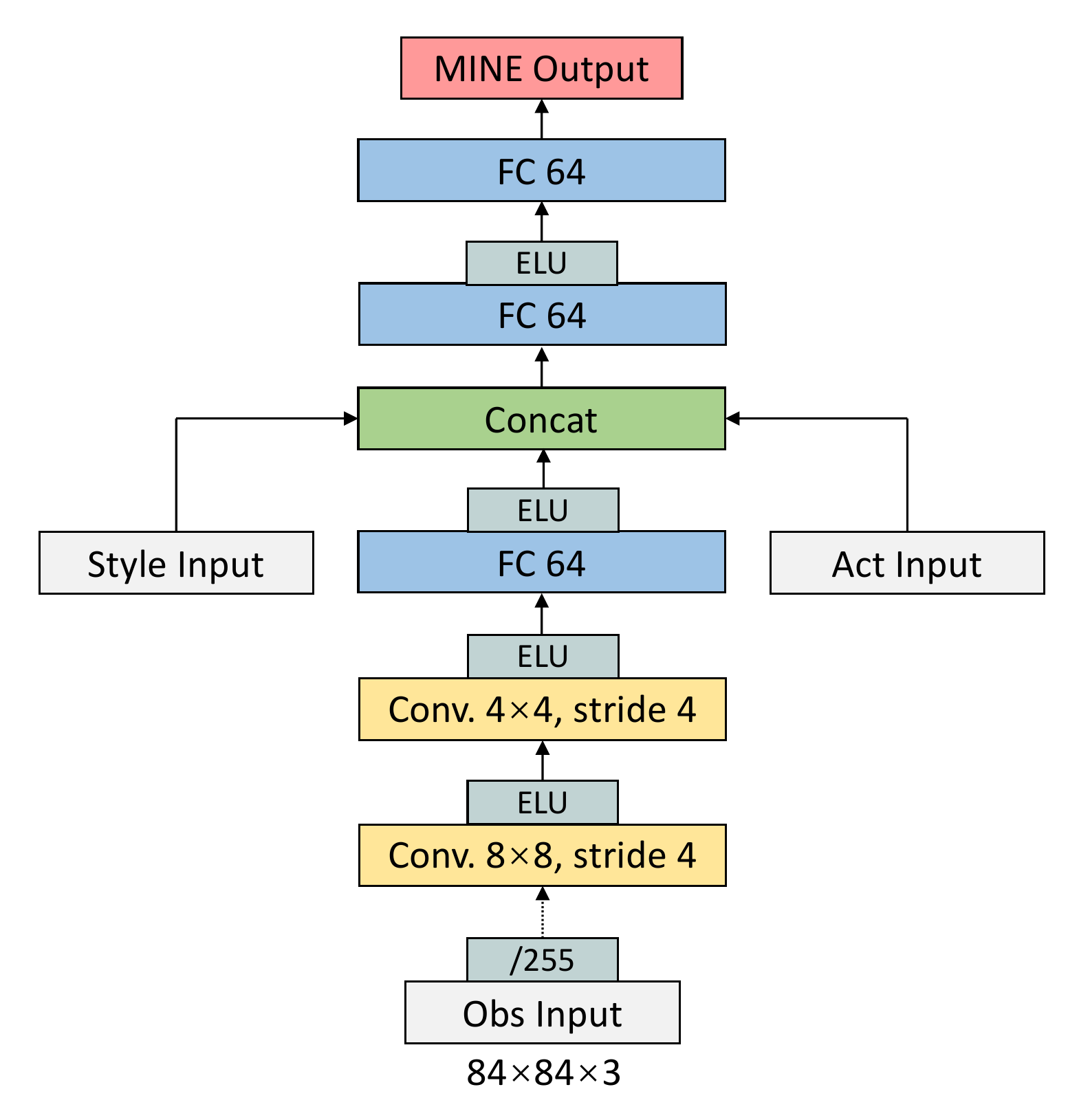}
\caption{Architecture of the MINE networks used in the Atari environment.}
\label{fig_net}
\end{figure}

Figure~\ref{fig_net} shows the MINE network structure, which is used in Atari tasks.

\subsection{Styles in SpaceInvaders Game}
\label{style_sp}

In the SpaceInvaders game, we employ two styles: the area style and the fire rate style.
The former divides the map into three distinct areas, distinguishing the agent’s preferences for moving toward each area.
The latter divides the agent's fire rate into three levels: (1) [0, 0.1); (2) [0.1, 0.3) and (3) [0.3, 1].



\end{document}